\documentclass{article} 
\usepackage{anonconf_conference,times}

\usepackage{amsmath,amsfonts,bm}

\def\eqref#1{equation~\ref{#1}}

\def\1{\bm{1}}

\DeclareMathAlphabet{\mathsfit}{\encodingdefault}{\sfdefault}{m}{sl}
\SetMathAlphabet{\mathsfit}{bold}{\encodingdefault}{\sfdefault}{bx}{n}

\usepackage{hyperref}
\usepackage{url}
\usepackage{graphicx}
\usepackage{amsmath}
\usepackage{amssymb}
\usepackage{mathtools}
\usepackage{amsthm}
\usepackage{multirow}
\usepackage{algorithm}
\usepackage{algpseudocode}
\usepackage[dvipsnames]{xcolor}
\usepackage{booktabs}
\usepackage{subcaption}
\usepackage{threeparttable}

\usepackage{titletoc}
\usepackage{animate} 

\newenvironment{packed_itemize}{
\begin{list}{\labelitemi}{\leftmargin=2em}
 \setlength{\itemsep}{0pt}
 \setlength{\parskip}{0pt}
 \setlength{\parsep}{0pt}
}{\end{list}}

\newtheorem*{prop}{Proposition}

\title{\textit{Patronus}: Interpretable Diffusion Models with Prototypes}

\author{Nina Weng, Aasa Feragen, Siavash Bigdeli
\\
  {Technical University of Denmark} \\
  \texttt{\small \{ninwe, afhar, sarbi\}@dtu.dk} \\
}

\anonconffinalcopy 
\begin{document}

\maketitle

\begin{abstract}

Uncovering the opacity of diffusion-based generative models is urgently needed, as their applications continue to expand while their underlying procedures largely remain a black box. 
With a critical question -- how can the diffusion generation process be interpreted and understood? -- we proposed \textit{Patronus}, an interpretable diffusion model that incorporates a prototypical network to encode semantics in visual patches, revealing \textit{what} visual patterns are modeled and \textit{where} and \textit{when} they emerge throughout denoising.
This interpretability of \textit{Patronus} provides deeper insights into the generative mechanism, enabling the detection of shortcut learning via unwanted correlations and the tracing of semantic emergence across timesteps. We evaluate \textit{Patronus} on four natural image datasets and one medical imaging dataset, demonstrating both faithful interpretability and strong generative performance. With this work, we open new avenues for understanding and steering diffusion models through prototype-based interpretability.\\
Our code is available at \href{https://github.com/nina-weng/patronus}{https://github.com/nina-weng/patronus}.

\end{abstract}    
\section{Introduction}

The generative capabilities of modern machine learning models, particularly diffusion models, have advanced significantly, enabling the creation of highly realistic samples that closely resemble real-world data. 
However, their opacity raises critical concerns, including bias amplification~\citep{luccioni2023stable}, unsafe content~\citep{qu2023unsafe}, and copyright violations~\citep{vyas2023provable}. 
Their lack of transparency makes it difficult to detect and mitigate these risks, highlighting a fundamental question: 
\textit{How can the diffusion generation process be interpreted and understood? }
Specifically, \textbf{what} visual patterns emerge during generation, \textbf{where} and \textbf{when} they appear, and to what extent they can be \textbf{controlled}. 
Addressing these questions is essential, not only for improving generative ability but also for ensuring interpretability, transparency, and ethical deployment, aligning with regulatory frameworks such as the EU AI Act.

Existing approaches to improving interpretability in diffusion-based visual generation typically fall into two categories. The first relies on post-hoc analysis to investigate how semantic information are encoded in  intermediate representations~\citep{kwon2022diffusion,lee2023diffusion,park2023understanding,haas2024discovering}. However, this method is inherently retrospective and limited in its ability to provide direct control over generation. The second approach introduces additional encoder-based semantic vectors for diffusion guidance~\citep{preechakul2022diffusion,leng2023diffusegae,wang2023infodiffusion}, which improves controllability, but often resulting in representations that are difficult to interpret. Moreover, these methods tend to capture global (e.g. face shape, pose) rather than fine-grained patterns (e.g. hair/make-up details, facial expressions), 
and the latter are crucial for interpretability.

\begin{figure}[t]
\centering
  \includegraphics[width=1.0\linewidth]{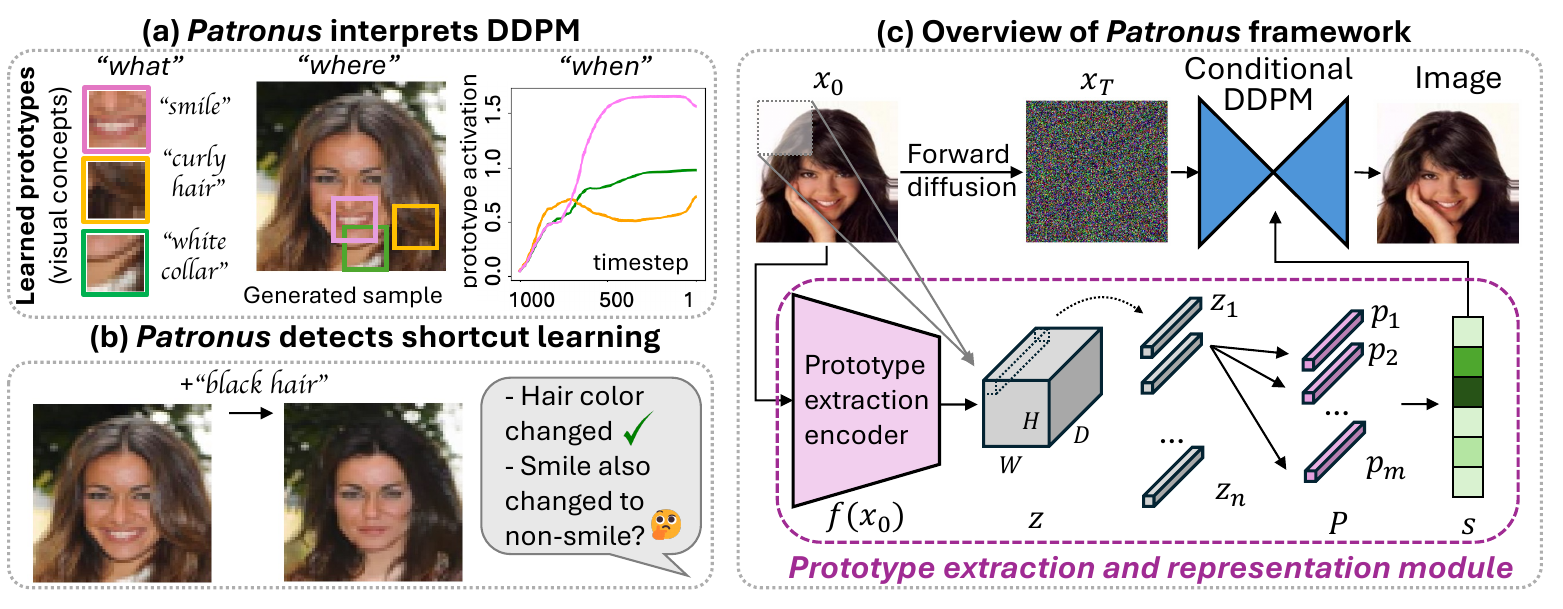}
\caption{\textbf{Proposed \textit{Patronus} model}. 
\textbf{(a) Interpretability}: By integrating a prototypical network as the encoder, \textit{Patronus} learns semantic prototypes (\textit{``what"}) and explains the generative process by revealing \textit{where} and \textit{when} they emerge.
\textbf{(b) Diagnosis:} 
Patronus could detect unwanted correlation (e.g., in this case hair color and smile) learned from training data.
\textbf{(c) Overview of \textit{Patronus}}: contains a prototypical network for prototype extraction and a conditional DDPM for generation.
}
    \label{fig:catchy_patronus}
\end{figure}

To address these limitations, we aim to develop a method that fulfills two key objectives:
(1) embedding interpretability directly into the model architecture, 
thereby providing \textbf{intrinsic transparency} and eliminating the need for post-hoc analysis of high-dimensional latent features;
(2) moving interpretability a step further from global to \textbf{local semantic meanings} and enabling the controllability. 

In this work, 
we propose \textit{Patronus}, an interpretable diffusion-based generative model integrated with a prototypical network for semantic encoding (Fig.~\ref{fig:catchy_patronus}). 
Our approach trains prototype features to capture localized patterns within image patches, and uses prototype activation vector to encode the presence of semantic information for diffusion guidance. This design effectively reduces the latent dimensionality while preserving sufficient semantic information. 
\textit{Patronus} allows directly visualizing the semantics of the learned prototypes by manipulating the condition signals on specific prototypes, thereby enhancing interpretability and enabling the diagnosis of shortcut learning.
Beyond that, \textit{Patronus} further reveals when prototypes emerge during the denoising process, 
providing temporal insights and guiding more efficient editing.

\paragraph{Our contributions are summarized as follows:}
\begin{packed_itemize}
    \item We propose  \textit{Patronus}: An interpretable diffusion model for image generation which incorporates a prototypical network for prototype learning and representation, alongside a conditional guidance by a prototype activation vector -- entirely without extra annotations.
    \item We introduce a novel method for visualizing learned prototypes, revealing the semantic meanings of them and how they engaged in the generation process. 
    \item Through extensive experiments, we show that \textit{Patronus} effectively captures semantically meaningful features within images, and achieves competitive latent quality and generation quality compared to SOTA methods.
    \item We empirically demonstrate the potential of \textit{Patronus} to diagnose hidden bias by detecting shortcut learning, offering a valuable tool for mitigating biases in generative models and promoting fairness in their deployment.
\end{packed_itemize}

\section{Related Work}

\subsection{Prototype-based Interpretability}
Our work is closely related to and motivated by ProtoPNet~\citep{chen2019looks}, which builds interpretable deep learning frameworks by learning \textit{prototypes}, intermediate representations of visually similar patterns between training and inference images.

Under such a design, a key challenge is visualizing the learned prototypes. \citet{li2018deep} used a decoder to reconstruct prototypes, which resulted in blurry visualizations due to data ambiguity. ProtoPNet instead identified the closest matching encoded representations from the training set via distance comparisons. However, this approach cannot guarantee capturing the true representative mode of the distribution.
Follow-up works~\citep{donnelly2022deformable,wang2023learning,ming2019interpretable,ghosal2021multi} explore alternative prototype design but \textit{do not} enhance visualization.
We propose a new method for prototype visualization that allow finding the most probable visual representation of a prototype. 
Additionally, the aforementioned studies only focus on \textit{classification}, our work focus on learning and visualizing prototypes for \textit{generative} models. 

\subsection{Interpretable Diffusion Models}

Recent work on explaining diffusion models has focused on the following two directions.

\paragraph{Semantic Interpretation of Internal Features.}

Diffusion models were traditionally viewed as lacking internal representations compared to other generative models, like VAE~\citep{kingma2013auto} and GAN~\citep{goodfellow2014generative}. However, ~\citet{kwon2022diffusion} pointed out that \textit{diffusion models possess a semantic latent space} within the U-Net's intermediate layer. They leverage the bottleneck for semantic control over the denoising process and propose the asymmetric reverse process to edit the image based on the discovered semantic meaning.  

Building on this idea, researchers have sought to interpret the latent space of diffusion models by grouping the latent vectors with different noise schedules~\citep{lee2023diffusion}, applying pullback metrics to obtain meaningful local latent basis~\citep{park2023understanding}, and uncovering semantically meaningful directions by both PCA and linear properties of the semantic latent space~\citep{haas2024discovering}. Other works~\citep{si2024freeu,tumanyan2023plug} extract structural information from the latent space and use it for I2I translation or prompted generation tasks. 

This line of work focuses on post-hoc analyses of existing diffusion models, whereas our approach provides \textit{intrinsic} interpretability by design. Moreover, while prior studies often rely on annotated data or external classifiers to uncover the semantic latent space, \textit{Patronus} learns prototypes in a fully \textit{unsupervised} manner.

\paragraph{Autoencoder-based semantic feature extraction for guidance.}
An alternative way to enhance the interpretability of diffusion models  is to integrate an additional encoder to extract semantic features for guidance. 
DiffAE~\citep{preechakul2022diffusion} pioneered this idea using a learnable encoder for latent semantics.
Expanding on this,
DiffuseGAE~\citep{leng2023diffusegae} proposed a group-supervised AutoEncoder to achieve better latent disentanglement, while 
InfoDiffusion~\citep{wang2023infodiffusion} reduced latent dimensionality and enforced mutual information constraints for more effective learning.

While these methods extract global semantic features, \textit{Patronus} instead captures local features through a prototypical network. Another key distinction lies in how the diffusion model is guided: Rather than direct semantic information, we use the \textit{prototype activation vector}.
This approach significantly reduces the dimensionality needed for diffusion guidance while preserving enough capacity in prototype features to encode the same semantic information.

\section{Patronus: Interpretable Diffusion Model}

Our proposed model, \textbf{Patronus:  Prototype-Assisted Transparent Diffusion Model}, is illustrated in Fig.~\ref{fig:catchy_patronus}c.
Designed to enhance transparency and interpretability in diffusion models, \textit{Patronus} incorporates a prototype extraction and representation module (Fig.~\ref{fig:catchy_patronus}c-bottom). This module learns patch-based prototypes within the image and computes similarity score for each prototype, which are then used to condition the diffusion process (Fig.~\ref{fig:catchy_patronus}c-top). 
We elaborate on the details of the prototype extraction and representation module in Sec.~\ref{sec:prototype_module} and the conditional DDPM in Sec.~\ref{sec:conditional_ddpm}. Furthermore, we explain how transparency and interpretability are achieved, including sampling strategy and manipulations, in Sec.~\ref{sec:method_interpretability}. We detail the unconditional sampling strategies in Sec.~\ref{sec:sampling_strategy}, and furthermore prove that adding conditions does not degrade the denoiser training in Sec.~\ref{sec:condition_benifits}.

\subsection{Prototype Extraction and Representation} \label{sec:prototype_module}

This module, inspired by ProtoPNet, consists of two key components:
The \textbf{prototype encoder} transforms input images into patch-based feature representations.
Utilizing the properties of convolutional neural networks (CNNs), each output neuron in the feature map corresponds to a specific patch of the input image, determined by the network's receptive field. 
This patch-based representation enables the model to focus on localized patterns and learn fine-grained prototypes.
The \textbf{activation vector} is derived by calculating similarity scores for each learned prototype, based on the distance between encoded patches and prototypes, where higher scores indicate stronger matches.

This module works as follows: 
As shown in Fig.~\ref{fig:catchy_patronus}c, given an input image $x_0$, the prototype encoder $f$ extracts features $z = f(x_0)$ into a tensor of shape $H \times W \times D$. In this work, the encoder is a 4-layer Conv–ReLU encoder.
The network learns $m$ prototypes in the latent feature space during training, denoted as $P = \{p_j\}_{j=1}^m$, each with the shape $1 \times 1 \times D$\footnote{whose generalization to $H_1 \times W_1 \times D$ as in~\citet{chen2019looks} is straightforward.}.
Each prototype $p_j$ can be interpreted as a latent encoding of a patch in the original pixel space. 
This patch, importantly, need not exist in the dataset but should lie within the plausible data distribution.

For the encoder output $z = f(x_0)$, each spatial region within $z$ that corresponds to the same size as a prototype ($1\times 1 \times D$) can be interpreted as representing a patch of $x_0$. Thus, $z$ can be decomposed into smaller regions as follows:
$z = \{z_i\}_{i=0}^{n}$,
where $n$ denotes the total number of patches encoded in $z$. In our case, $n = H \times W$.

To calculate the similarity between the encoded features $z$ and the learned prototypes $P$, we begin by computing the squared $L2$ distance between each spatial feature $z_i$ and each prototype $p_j$: $d^2(z_i,p_j) = ||z_i-p_j ||^2$. Next, the minimum distance across the spatial dimensions is selected for each prototype:
$d^2_{min,j} = max(-d^2_j,$ kernel size $=(H,W))$, where $d^2_j$ is the set of distances for the $j_{th}$ prototype across all spatial positions, and the kernel size aligns with the feature map's spatial dimensions $H$ and $W$.
The sequence of minimum distances for all prototypes is converted into an activation vector $s$ using a log transformation: 
$s = \log (\frac{d^2 +1}{d^2+\epsilon})$
, where $\epsilon$ is a small positive constant.

\subsection{Conditional Diffusion Process} \label{sec:conditional_ddpm}

Denoising diffusion probabilistic models (DDPM) form a class of generative models that learn data distributions by iteratively denoising a noisy latent representation. The process involves \textbf{(1) a forward diffusion process}, 
where Gaussian noise is progressively added to a data sample $x_0$ over $T$ timesteps, producing noisy latents $x_t$, defined as 
\begin{equation}
    q(x_t \mid x_{t-1}) = \mathcal{N}(x_t; \sqrt{\alpha_t} x_{t-1}, (1 - \alpha_t) \mathbf{I}).
\end{equation}
Here, the marginal distribution of $x_t$ given $x_0$ is:
\begin{equation}
q(x_t \mid x_0) = \mathcal{N}(x_t; \sqrt{\bar{\alpha}_t} x_0, (1 - \bar{\alpha}_t) \mathbf{I}),
\end{equation}
where $\bar{\alpha}_t = \prod_{i=1}^t \alpha_i$ and $\alpha_t = 1-\beta_t$, where $\beta_t$ is the variance of the Gaussian noise added at $t$. 
Furthermore, DDPMs rely on \textbf{(2) a reverse generative process} that removes noise given $t$:
\begin{equation}
    p_\theta(x_{t-1} \mid x_t) = \mathcal{N}(x_{t-1}; \mu_\theta(x_t, t), \Sigma_\theta(x_t, t)).
\end{equation}
To enable \textbf{conditional generation}, we modify the reverse process to be conditioned on the prototype activation vector $s$. Therefore the updated reverse process is: 
\begin{equation}
    p_\theta(x_{t-1} \mid x_t, s) = \mathcal{N}(x_{t-1}; \mu_\theta(x_t, t,s), \Sigma_\theta(x_t, t)).
\end{equation}
The training objective remains based on the standard noise-prediction loss used in DDPM. For a given noisy sample $x_t$, timestep $t$ and noise $\epsilon$ the model minimizes the loss:
\begin{equation}
    \mathcal{L}_{ddpm} = \mathbb{E}_{x_0,\theta,t} [||\epsilon - \epsilon_\theta(x_t,t,s) ||^2],
\end{equation}
where $\epsilon_\theta$ is the learned denoiser. As the loss indicates, our guidance does not change the model's output -- it only encourages its reasoning to utilize prototypes for transparency.

\subsection{Transparency and Interpretability of Patronus} \label{sec:method_interpretability}

The similarity score $s_j$ quantifies the activation of the $j_{th}$ prototype in a given input, indicating the presence of specific semantic patterns. The model thus conditionally generates samples guided by interpretable semantic information.

\paragraph{Visualizing learned prototypes.} \label{sec:vis_proto}

Integrating a prototypical network as a semantic meaning extraction module brings inherent interpretability: Each learned prototype vector $p_j$ represents a patch in the image domain.  
In ProtoPNet, those patches are retrieved by greedily searching all candidate patches in the training set for the closest embeddings.

We argue that the learned prototypes do not need to correspond directly to specific training patches but should instead align with the overall distribution of the training data.  
To support this, we propose a novel prototype visualization method with the following steps:
\begin{enumerate}
    \item Compute the activation vector $s = \{s_j\}_{j=1}^m$, for a given sample $x_0$, where $s_j$ represents the similarity score between $x_0$ and the $j_{th}$ prototype. 
    \item For target prototype $J$, increase its similarity score $s_J$ to the plausible maximum while keeping all other scores unchanged.
    The updated activation vector
    $s' = \{s'_j\}_{j=1}^m$ is defined as:
$        s'_j =
\begin{cases}
    s_j, & \text{if } j \neq J \\
    max(s_X), & \text{if } j = J  
\end{cases} $.
    Here, $s_X$ represents similarity scores from a representative subset, constraining $s_J$ within a plausible range.
    \item Using the updated activation vector $s'$ to sample a new image $x'$ conditioned on $s'$. 
    \item Identify the most activated patch $x'_i$ in $x'$ that corresponds to the target prototype $J$. This patch $x'_i$ serves as the visual representation of $p_J$.
\end{enumerate}
\noindent This method could also be used to visualize prototypes in other prototypical deep learning models.

\paragraph{Manipulation using the prototype activation vector.}
Manipulating images is a natural downstream task for \textit{Patronus}, as adjusting a specific prototype similarity score $s_j$ and conditionally generating a new sample allows us to effectively and semantically control the image content:
\begin{equation}
    p_\theta(x_{t-1} \mid x_t, s') = \mathcal{N}(x_{t-1}; \mu_\theta(x_t, t,s'), \Sigma_\theta(x_t, t))
\end{equation}

\paragraph{Deterministic reverse process via DDIM sampling.}

Both the visualization of prototypes and their manipulation via activation vectors build on the DDPM sampling process. However, for stricter control over the stochasticity introduced by random noise, the Denoising Diffusion Implicit Models (DDIM) sampler provides an alternative approach. Here, the denoiser is given by 
    $p_\theta(x_{t-1} \mid x_t, s) = \mathcal{N}(x_{t-1};\mu_t, \sigma^2 \mathbf{I})$, with mean
    $\mu_t = \sqrt{\bar{\alpha}_{t-1}} \cdot \hat{x}_0 + \sqrt{1-\bar{\alpha}_{t-1} -\sigma^2} \cdot {\epsilon}_\theta(x_t,t,s)$ and variance
    $\sigma^2 = \eta^2 \cdot \frac{1-\bar{\alpha}_{t-1}}{1-\bar{\alpha}_{t}} \cdot (1 - \frac{\bar{\alpha}_t}{\bar{\alpha}_{t-1}})$.
By setting $\eta = 0.0$, the process becomes deterministic, leading to the following update for the reverse process: 
    $x_{t-1} = \sqrt{\bar{\alpha}_{t-1}} \cdot \hat{x}_0 + \sqrt{1-\bar{\alpha}_{t-1} } \cdot {\epsilon}_\theta(x_t,t,s)$.
Here, $\hat{x}_0$ is the estimated denoised image, computed as $
    \hat{x}_0 = \frac{1}{\sqrt{\bar{\alpha}_t}}(x_t - \sqrt{1-\bar{\alpha}_{t}} \cdot {\epsilon}_\theta(x_t,t,s))$.

This DDIM sampling requires $x_T$, which represents the initial noise in the diffusion process. This could be obtained by performing a deterministic backward generative process, given by
    $x_{t+1} = \sqrt{\bar{\alpha}_{t+1}} \cdot \hat{x}_0 + \sqrt{1-\bar{\alpha}_{t+1}} \cdot {\epsilon}_\theta(x_t,t,s)$.

\subsection{Unconditional Sampling Strategy} \label{sec:sampling_strategy}
For unconditional sampling, we train an auxiliary latent diffusion model $p(s_{t-1}|s_t,t)$ to sample $s$. 
During training, we first jointly optimize the prototypical encoder with the conditional DDPM; subsequently, the latent diffusion model is trained with the parameters of prototype encoder frozen.

\subsection{Adding the Condition to the Objective} \label{sec:condition_benifits}
In \textit{Patronus}, the prototype encoder is jointly optimized with the denoiser. 
To show that this simultaneous training does not degrade the generated distribution, we analyze how updating the condition $s$ affects the likelihood.
For ease of derivation, we use the Evidence Lower Bound (ELBO) as an equivalent objective to generalize denoising losses~\citep{ho2020denoising}.

\begin{prop}
Any ELBO-improving update for the encoder always leads to progress. Such an update either increases the conditional likelihood of the data under the model or reduces the KL divergence between the generated distribution and the underlying data distribution.
\end{prop}
\begin{proof}
Let $z = x_{1:T}$ denote the forward noised latents generated from a fixed
forward process $q(z \mid x)$, as is standard in diffusion models.
For parameters $s$, define the per-sample evidence lower bound (ELBO)
\[
\mathrm{ELBO}(x; s)
=
\mathbb{E}_{q(z\mid x)}
\!\left[
    \log p_\theta(x,z \mid s)
    -
    \log q(z\mid x)
\right],
\]
and the conditional log-likelihood
\[
\log p_\theta(x \mid s)
=
\mathrm{ELBO}(x; s)
+
\mathrm{KL}\!\left(
    q(z\mid x)\,\big\|\,p_\theta(z\mid x,s)
\right).
\]
Consider an update $s^i \mapsto s^{i+1}$ that satisfies
$\mathrm{ELBO}(x; s^{i+1}) \ge \mathrm{ELBO}(x; s^{i})$ for the data point $x$.
Then the following identity holds:
\begin{equation}
\label{eq:elbo-likelihood-diff}
\log p_\theta(x \mid s^{i+1}) - \log p_\theta(x \mid s^i)
=
\big[
    \mathrm{ELBO}(x; s^{i+1})
    -
    \mathrm{ELBO}(x; s^{i})
\big]
+
\Delta \mathrm{KL}(x),
\end{equation}
where
$
\Delta\mathrm{KL}(x)
=
\mathrm{KL}\!\left(q(z\mid x)\,\|\,p_\theta(z\mid x,s^{i+1})\right)
-
\mathrm{KL}\!\left(q(z\mid x)\,\|\,p_\theta(z\mid x,s^{i})\right).
$

Given that the $\mathrm{ELBO}$ difference term is positive, Equation~\ref{eq:elbo-likelihood-diff} implies that one of the following
(improving) outcomes must occur:

\textbf{Conditional likelihood improves.}
If
$\Delta\mathrm{KL}(x) \le 0$, then
\[
\mathrm{KL}\!\left(q(z\mid x)\,\|\,p_\theta(z\mid x,s^{i+1})\right)
<=
\mathrm{KL}\!\left(q(z\mid x)\,\|\,p_\theta(z\mid x,s^{i})\right).
\]
meaning that the model posterior $p_\theta(z\mid x,s^{i+1})$ more closely matches the forward distribution $q(z\mid x)$.  
This reduces the gap between the ELBO and the true conditional log-likelihood, yielding a tighter variational bound.

\textbf{Generated samples become more probable.}
If
$\Delta\mathrm{KL}(x) > 0$ and considering that ELBO is also increased, then necessarily
\[
\log p_\theta(x \mid s^{i+1})
\ge
\log p_\theta(x \mid s^{i}),
\]
i.e.\ the model assigns higher probability to~$x$.

Hence, any update of the encoder that increases the ELBO either increases the conditional likelihood of the data or reduces the KL divergence between the model posterior and the forward process, guaranteeing progress in approximating the data distribution.
\end{proof}

This analysis is inapplicable under alternative training objectives for latent condition (e.g., InfoDiff).

\section{Experiments and Results}

In our experiments, we first assess the semantic meaning of learned prototypes (Sec.~\ref{sec:semantic_meaning}) through reconstruction, interpolation, extrapolation, and manipulation tasks. We also investigate prototype visualization and their consistency.
We then analyze prototype quality and generation performance (Sec.~\ref{sec:p_quality},~\ref{sec:gen_quality}). Finally, we explore how \textit{Patronus} aids in diagnosing shortcut learning in diffusion models by identifying potentially unwanted correlations via prototype similarity scores (Sec.~\ref{sec:res_diagnosis}).

We use five datasets: Quantitative evaluation on FMNIST~\citep{xiao2017fashion}, CIFAR-10~\citep{krizhevsky2009learning},  FFHQ~\citep{karras2019style}; qualitative analysis on CheXpert~\citep{irvin2019chexpert}; and in-depth quantitative and qualitative experiments
on CelebA~\citep{liu2015faceattributes}.
We use 100 prototypes for all datasets except FMNIST, 
which has 30. Prototypes are encoded with a shape of (1,1,128).

\subsection{Semantic Meaning of the Learned Prototypes.} \label{sec:semantic_meaning}

\begin{figure}[t]
\centering
  \includegraphics[width=1.0\linewidth]{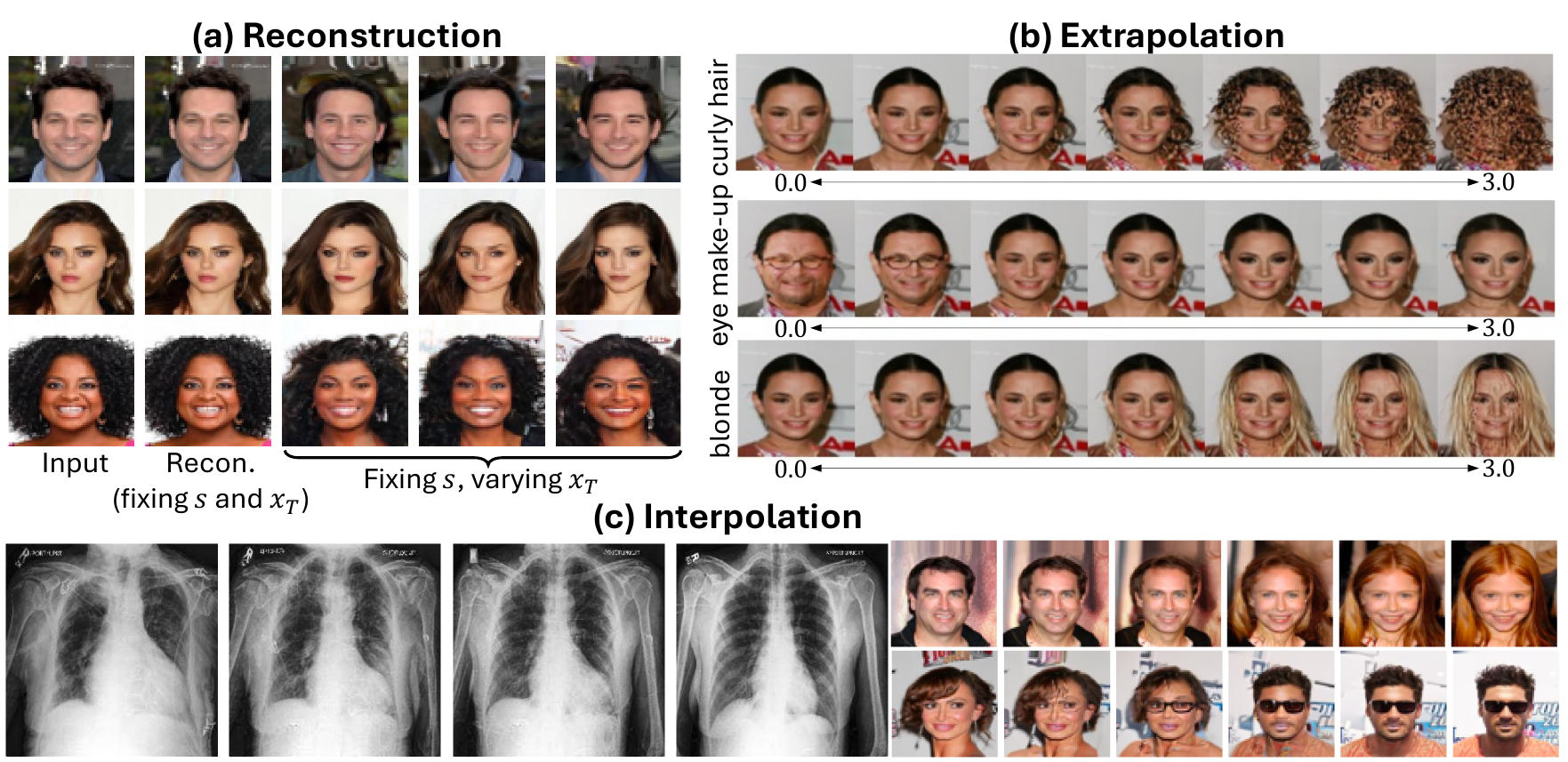}
\caption{\textbf{Assessing the semantic meaning of the learned prototypes.}
(a) Reconstruction.
(b) Extrapolation.
(c) Interpolation. Left: CheXpert, from 75-year-old female $w/o$ enlarged heart (left) to 27-year-old male $w/$ enlarged heart (right). Right: 2 examples from CelebA.
}
    \label{fig:semantic_meaning}
\end{figure}

We firstly verify that the learned prototypes are \textit{semantically meaningful} by the following analyses.
\paragraph{Reconstructing image semantics from prototype activations.} \label{sec:letent_representation}
We extract the prototype activation vector $s = Enc(x_0)$ and sample random noise $x_T \sim \mathcal{N}(0, \mathbf{I})$ to generate new images $\hat{x}(s, x_T)$. Fig.~\ref{fig:semantic_meaning}a presents one reconstructed image (using the same $x_T$) with three variations (using random $x_T$). 
The results show that the majority of semantic information in the images is accurately recreated, confirming that the prototypes effectively capture meaningful semantic features.

\paragraph{Interpolation.} \label{}

Given two images $x_0^{1}$ and  $x_0^{2}$, we first retrieve its corresponding prototype activation vector and starting noise by reverse DDIM process: $(s^1,x_T^1)$ and $(s^2,x_T^2)$,  and then generate new samples using $(\mathrm{Lerp}(s^1,s^2;t),\mathrm{Slerp}(x_T^1,x_T^2;t))$ for steps $t\in[0,1]$, where $\mathrm{Lerp}$/$\mathrm{Slerp}$ represents linear/spherical linear interpolation respectively. Results are shown in Fig.~\ref{fig:semantic_meaning}c.

\begin{figure}[t]
\centering
  \includegraphics[width=1.0\linewidth]{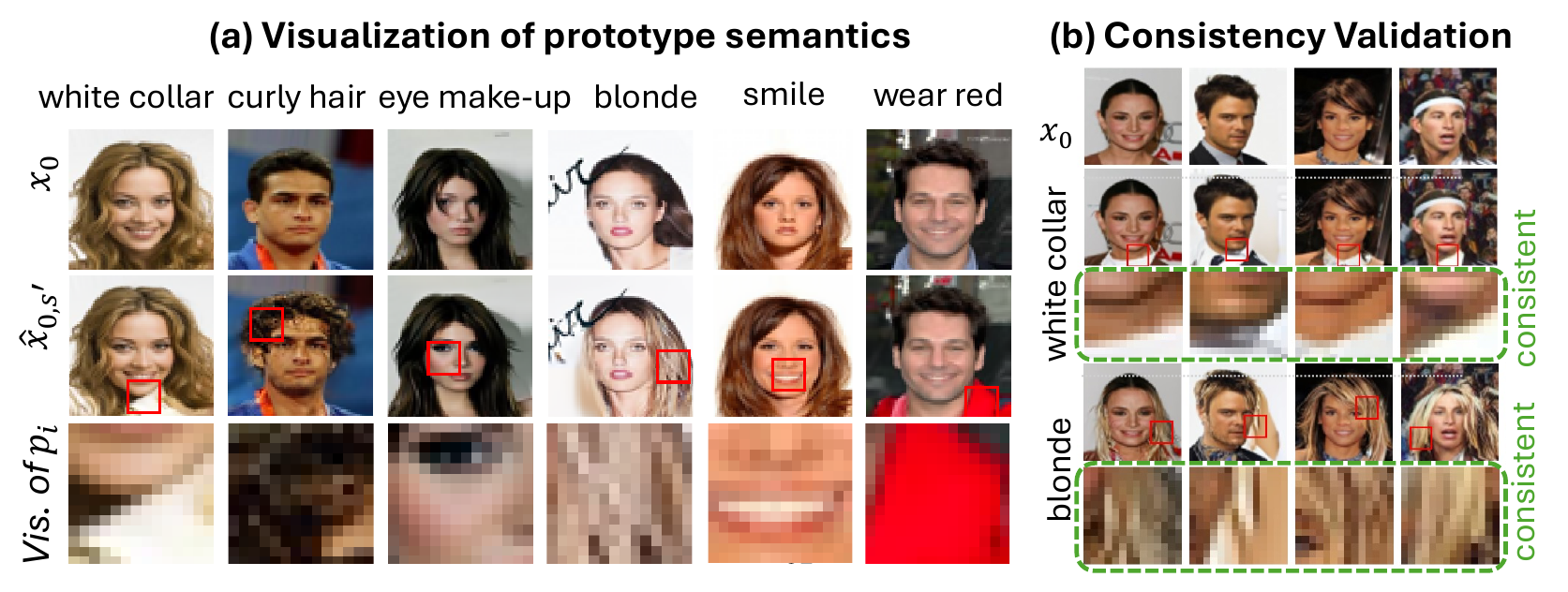}
\caption{\textbf{Prototype visualization and consistency.} 
(a) Prototype visualization.
Here, $x_0$ denotes the original image, $\hat{x}_{0,s'}$ denotes the generated image guided by condition $s'$, where $s'$ is the enhanced prototype activation on $j$-th prototype. 
Red square highlights the most activated patch,
which is considered as the visual representation of the chosen prototype, also shown in the third row. Note that prototype semantics are not pre-annotated but inferred through observation.
(b) Visualization across random samples demonstrates that each prototype encodes consistent semantics.
}
    \label{fig:vis_proto}
\end{figure}

\paragraph{Visualization of the learned prototypes and consistency validation.}\label{sec:res_vis_proto}
Unlike other autoencoder-based diffusion models, our approach is explicitly designed to yield an interpretable semantic latent space, where each prototype is trained to capture distinct semantic content.
In practice, we realize this by amplifying a prototype’s activation and extracting the most responsive patch (see Sec.~\ref{sec:vis_proto}). Fig.~\ref{fig:vis_proto}a shows selected prototypes with their semantic visualizations, and Fig.~\ref{fig:vis_proto}b demonstrates that the most activated patches remain semantically consistent across samples.

\paragraph{Manipulation and extrapolation.}
By adjusting the condition $s$ we can edit the image with specific semantic requests. 
Furthermore, pushing a selected dimension of $s$ to extreme values (ranging from 0.0 to 3.0) results in a smooth and continuous enhancement of the associated semantic information, as demonstrated in Fig.\ref{fig:semantic_meaning}b. Unlike interpolation, which remains within the observed range, this process extends beyond the original data distribution, making it an \textit{extrapolation}.

\subsection{Prototype Quality} \label{sec:p_quality}
\paragraph{Prototype capability in semantic representation.}

We test the prototype quality via a downstream classification task on the prototype activation vectors $s$ using a logistic regression classifier trained with 5-fold cross validation, reporting AUROC in Tab.~\ref{tab:p_and_gen_quality_celeba} \& \ref{tab:p_quality_other_ds}. Our model outperform 3 out of 4 datasets in latent (prototype) quality, with particularly strong performance on CelebA and FFHQ. 
The lower latent quality for FMNIST may stem from \textit{Patronus} prioritizing localized features, while DiffAE and InfoDiff emphasize global structures, which better capture the semantic information of FMNIST due to its high inter-class variability. For the datasets where semantic information is more localized, \textit{Patronus} achieves a marked improvement.

\paragraph{Prototype disentanglement.}

We quantify the disentanglement on CelebA using TAD~\citep{yeats2022nashae}. Following~\citet{wang2023infodiffusion}, we first remove the highly correlated attributes, then compute the AUROC score for each dimension of the prototype activation vector $s$. An attribute is ``captured" if any dimension achieves AUROC $>$ 0.75. TAD is the sum of AUROC differences between the top two predictive dimensions per captured attribute. As shown in Tab.~\ref{tab:p_and_gen_quality_celeba}, \textit{Patronus} outperforms previous models in both TAD and captured attributes.

\subsection{Generation Quality}\label{sec:gen_quality}

\begin{table}[t]
    \centering
    \caption{\textbf{Prototype quality and generation quality on CelebA.} }
    \label{tab:p_and_gen_quality_celeba}
    \footnotesize
    
\scalebox{0.95}{\begin{tabular}{l|cccc} \toprule
         & TAD 	$\uparrow$ & Attrs $\uparrow$ & Latent AUROC$\uparrow$ & FID $\downarrow$\\
         \midrule
         DiffAE& 0.16$\pm$0.01 & 2.0$\pm$0.0 & 0.80$\pm$0.00 & 22.7$\pm$2.1\\
         InfoDiff & 0.30$\pm$ 0.01 & 3.0$\pm$0.0 & 0.84$\pm$0.00 & 23.6$\pm$1.3 \\
         \;$w/$ learned $z$ & 0.30$\pm$ 0.01 & 3.0$\pm$0.0 & 0.84$\pm$0.00 & 22.3$\pm$1.2\\
        \midrule
        Patronus & \textbf{0.43$\pm$0.02} & \textbf{9.0$\pm$0.0} & \textbf{0.87$\pm$0.00} & 14.6$\pm$0.1\\
        \;$w/$ learned $s$ & \textbf{0.43$\pm$0.02} & \textbf{9.0$\pm$0.0} & \textbf{0.87$\pm$0.00} &
        \textbf{4.8$\pm$0.0} \\

         \bottomrule 
    \end{tabular}
}
    
\end{table}

\begin{table}[tb]
    \centering
    \caption{\textbf{Prototype quality and generation quality on FashionMNIST, CIFAR-10, and FFHQ.}}
    \label{tab:p_quality_other_ds}
    \footnotesize
    
\scalebox{0.88}{\begin{tabular}{l|cc|cc|cc} \toprule
            
         & \multicolumn{2}{c}{\textbf{FMNIST}} & \multicolumn{2}{|c}{\textbf{CIFAR-10}}& \multicolumn{2}{|c}{\textbf{FFHQ}} \\
         & Latent AUROC $\uparrow$& FID $\downarrow$ & Latent AUROC $\uparrow$&FID $\downarrow$& Latent AUROC $\uparrow$& FID $\downarrow$  \\
 
         \midrule
         DiffAE & 0.84$\pm$0.00 & 8.2$\pm$0.3 & 0.40$\pm$0.01 & 32.1$\pm$1.1 &0.61$\pm$0.00 & 31.6$\pm$1.2\\
         InfoDiff & \textbf{0.84$\pm$0.00}& 8.5$\pm$0.3 & 0.41$\pm$0.00 & 32.7$\pm$1.2 &0.61$\pm$0.00 & 31.2$\pm$1.6\\
         \;$w/$ learned $z$ & \textbf{0.84$\pm$0.00} & 7.4$\pm$0.2 & 0.41$\pm$0.00 & 31.5$\pm$1.8 &0.61$\pm$0.00& 30.9$\pm$2.5\\
        \midrule
        Patronus& 0.82$\pm$0.00 & 14.7$\pm$ 0.3 &\textbf{0.54$\pm$0.01} &32.9 $\pm$ 0.4 & \textbf{0.92$\pm$0.00}& 37.3$\pm$0.2\\
        \;$w/$ learned $s$ &0.82$\pm$0.00& \textbf{2.6 $\pm$ 0.1}& \textbf{0.54$\pm$0.01} &\textbf{8.0 $\pm$ 0.1}& \textbf{0.92$\pm$0.00}& \textbf{24.1 $\pm$ 0.1} \\

         \bottomrule 
    \end{tabular}
}
    
\end{table}

We assess generative quality using the Fréchet Inception Distance (FID), averaged over five random test sets of 10,000 images. Evaluation covers both unconditional and prototype-conditioned generation, where the latter incorporates learned prototype activation vectors from the test set (see Tab.\ref{tab:p_and_gen_quality_celeba} \& \ref{tab:p_quality_other_ds}).
\textit{Patronus} significantly outperforms previous methods in prototype-conditioned generation across all four datasets. In the unconditional setting, it achieves state-of-the-art performance on CelebA and remains competitive on others; noting that its effectiveness depends on the training quality of both \textit{Patronus} and the latent diffusion model.

\subsection{Diagnosing Shortcut Learning in Diffusion Models}\label{sec:res_diagnosis}

\begin{figure}
    \centering
    \includegraphics[width=1.0\linewidth]{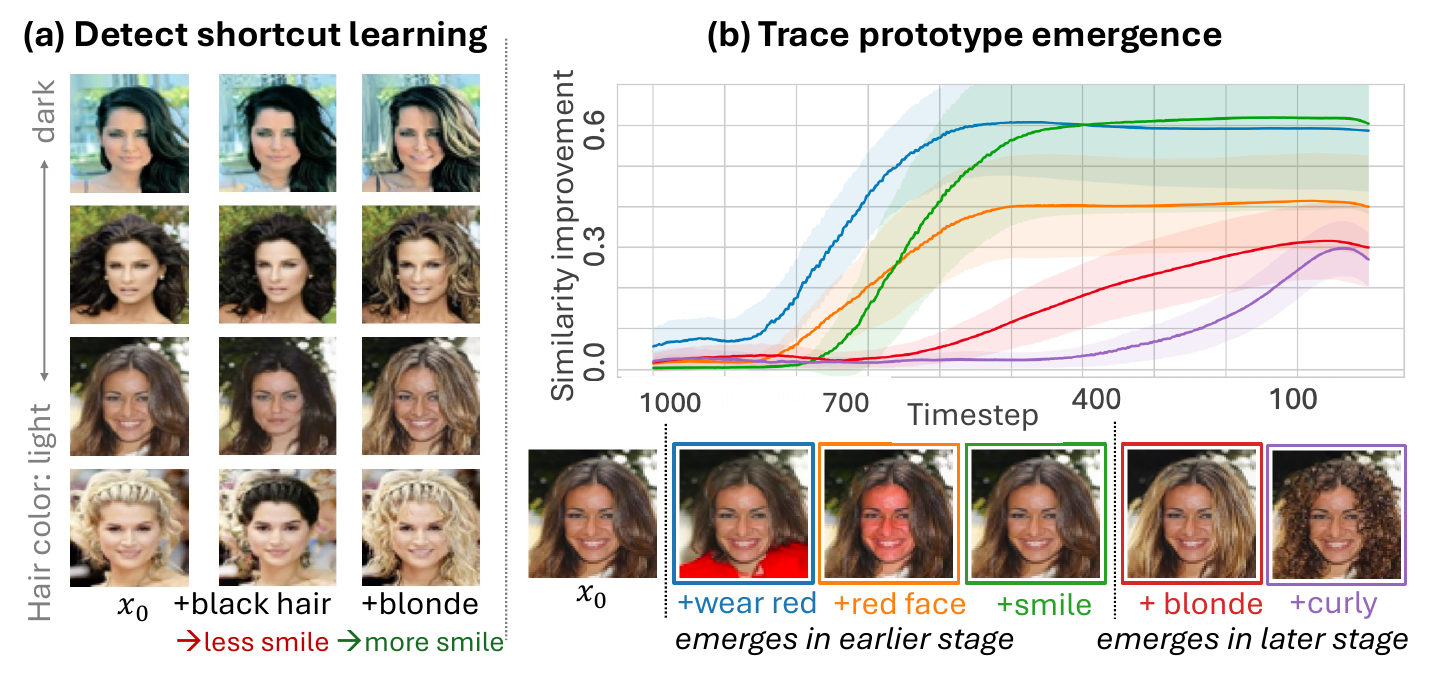}
    \caption{\textbf{\textit{Patronus} as an interpretable tool} for 
(a) \textbf{Detecting shortcut learning.} Enhancing hair-color prototypes reveals their correlation with other attributes (e.g., smile), thereby exposing unwanted biases. 
(b) \textbf{Tracing prototype emergence.} Different prototypes appear at different stages of the diffusion process, as indicated by similarity score improvements, which provides insights for more effective image editing strategies.}
    \label{fig:interpretable_tool}
\end{figure}

We manipulated a subset of CelebA to introduce an unwanted correlation between hair color and smile: all blonde/brown-haired images smile, while black-haired ones do not. 
Results confirm the model captures the introduced bias, as increasing the black hair prototype shifts the smile property from ``smile" to ``non-smile" and vice versa (see Fig.~\ref{fig:interpretable_tool}a). Similar patterns emerge without subset manipulation.  In Fig.~\ref{fig:semantic_meaning}b, decreasing ``eye makeup" reduces female features.
This emphasizes how \textit{Patronus} can be utilized to discover unwanted model behavior, such as shortcut learning in this case.

\section{Discussion and Conclusion} 
\label{sec:dicussion}

\paragraph{Prototypes Emerge at Different Times during Generation.}
Given a generated image, \textit{Patronus} reveals \textit{when} each prototype emerges in the generation process by obtaining the prototype similarity score from estimated $\hat{x}_0$ at each timestep (more details in Appendix), as shown in the top-right corner of Fig.~\ref{fig:catchy_patronus}a. Thus, by subtracting the two time-sequential $s$ of the semantically-enhanced image $\hat{x}_{0,s\prime}$ and $x_0$, we illustrate how each prototype emerges temporally in the diffusion process. 
As shown in Fig.~\ref{fig:interpretable_tool}b, none of the prototypes have a significant emergence in the first 200 stages of generation.
Interestingly, prototypes relating to lower spatial frequency attributes appear earlier during generation, such as ``wearing red"; while higher spatial frequency attributes, like ``curly hair", emerge later in the generation process. 
This insight could potentially improve the efficiency of image editing or counterfactual generation by guiding how far an image should be reversed in the diffusion process. It could also support bias mitigation by leveraging the same mechanism once unwanted correlations are detected.

\paragraph{Prototype Correlation and Collapse.}
We test whether correlation between prototypes is caused by \textit{prototype collapse}, where multiple prototypes represent the same semantics. To assess this, we introduce a Prototype Distinct Loss to encourage prototype disentanglement and evaluate its impact compared to using the denoiser loss alone. The Prototype Distinct Loss is defined as: $
\mathcal{L}_{distinct} = \frac{1}{N} \sum_{i=1}^{N} \max(0, \delta - \min_{j \neq i} D_{ij})$, where $D$ is the cosine distance with absolute similarity: $
D_{ij} = 1 - |\frac{p_i \cdot p_j}{\|p_i\| \|p_j\||}|$.
The margin $\delta$ is set to 0.5 and 1.0, where $\delta = 1.0$ enforces prototypes to be orthogonal. 
We initialize the new model using the original network parameters and train it for an additional 100 epochs using $L_{ddpm}+L_{distinct}$. 
Results show that the new models do not see a substantial change in the learned prototypes, suggesting that the prototypes optimized via the denoising objective are already sufficiently decorrelated without explicit regularization (see Appendix).

\paragraph{Do We Capture All Relevant Attributes?}
While \textit{Patronus} shows great ability in capturing attributes, we notice that 
global features, e.g. gender and age, are harder to find in one specific prototype. This could result from the patch-based prototypical encoder, making non-local features hard to capture. 
See Appendix for illustrative visual examples.

\subsection{Conclusion}

We propose \textit{Patronus}, an interpretable diffusion model integrated with a prototypical network. It enables intuitive interpretation of the generation process by visualizing learned prototypes (\textit{what}) and identifying \textit{where} and \textit{when} they appear. It also supports semantic manipulation through prototype activation vector.
Experiments show that \textit{Patronus} achieves competitive performance and learns meaningful prototype-based representations.
We further explore its capability to diagnose unwanted correlations in the generative process. We believe \textit{Patronus} offers valuable insights into interpretable diffusion models by bridging diffusion and prototypical networks.

\section*{Acknowledgements.} 
Work on this project was partially funded by DTU Compute, the Technical University of Denmark; the Pioneer Centre for AI (DNRF grant nr P1); and the Novo Nordisk Foundation through the Center for Basic Machine Learning Research in Life Science (MLLS, grant NNF20OC0062606). The funding agencies had no influence on the writing of the manuscript nor on the decision to submit it for publication.

\bibliography{reference}
\bibliographystyle{anonconf_conference}

\end{document}